\documentclass[pdflatex,sn-mathphys-num]{sn-jnl}

\usepackage{graphicx}%
\usepackage{multirow}%
\usepackage{amsmath,amssymb,amsfonts}%
\usepackage{amsthm}%
\usepackage{mathrsfs}%
\usepackage[title]{appendix}%
\usepackage{xcolor}%
\usepackage{textcomp}%
\usepackage{manyfoot}%
\usepackage{booktabs}%
\usepackage{makecell}
\usepackage{tcolorbox}
\usepackage{algorithm}%
\usepackage{algorithmicx}%
\usepackage{algpseudocode}%
\usepackage{listings}%
\usepackage{enumitem}

\theoremstyle{thmstyleone}%
\theoremstyle{thmstyletwo}%

\theoremstyle{thmstylethree}%

\begin{document}

\title[Article Title]{MLLM‑Routed Heterogeneous Ensembles for Robust Cross‑Dataset Image Classification}


\author*[1,2]{\fnm{Daniel} \sur{Perkins}}\email{dperki16@vols.utk.edu}

\author[1]{\fnm{John} \sur{Squires}}\email{jyv397@vols.utk.edu}

\author[1,2]{\fnm{Janou} \sur{Milligan}}\email{jmillig5@vols.utk.edu}

\author[1]{\fnm{Chandra} \sur{Raskoti}}\email{craskoti@vols.utk.edu}

\author[1,2]{\fnm{Linda} \sur{Ungerboeck}}\email{lungerbo@vols.utk.edu}

\affil[1]{\orgname{University of Tennessee}, \city{Knoxville}, \state{Tennessee}, \country{USA}}

\affil[2]{\orgname{The Bredesen Center for Interdisciplinary Research and Graduate Education}}


\abstract{Modern image classification models excel when trained on single task‑specific datasets but often struggle to generalize across domains and difficulty levels. We propose ARMDIL, an Adaptive Router for Multi-Domain Image Classification with LLMs. ARMDIL is an ensemble that uses a multimodal large language model (MLLM) agent to dynamically route each image to the most suitable vision backbone. Our diverse ensemble employs convolutional neural networks (ResNets), self-supervised representation learners (SSL), and vision language models (VLMs), each trained on a unified label space constructed from multiple image datasets with differing distributions and characteristics. Empirical evaluations illuminate the distinct capabilities and vulnerabilities of each architecture across disparate visual domains. Crucially, we show that ARMDIL effectively navigates these tradeoffs, performing competitively with specialized training-based routers. Furthermore, it drastically improves adaptability by allowing new information to be integrated via simple prompt modifications, while enhancing interpretability through natural language reasoning traces. These advances in cross-dataset image classification pave the way for more reliable general-purpose vision systems such as AI assistants and autonomous robots.}




\maketitle

\section{Introduction}
\label{sec:intro}

Modern image classification systems are increasingly deployed across a vast spectrum of real-world environments, from medical imaging and agriculture to facial recognition and autonomous driving. For instance, while classification models applied to MRI scans assist clinicians in diagnosing brain tumors, those embedded in autonomous vehicles are crucial for detecting road hazards. Although contemporary models achieve remarkable accuracy within these specific, isolated domains, the sheer diversity of visual problems presents a critical challenge: a model highly optimized for one task frequently fails on another. The structural nuances of a chest X-ray, for example, share very few characteristics with the object-centric nature of natural images.

Convolutional Neural Networks (CNNs)~\cite{OShea2015AnIT} and Vision Transformers (ViTs)~\cite{dosovitskiy2021an} were the first models to deliver highly successful results on single tasks, replacing fully connected linear layers with operations that model spatial relationships. While these standard deep learning architectures excel within their training distributions, adapting them to new tasks or evolving classification goals requires extensive retraining. This constant need for massive labeled datasets and computational resources limits their robustness and applicability in dynamic, real-world scenarios.

Self-Supervised Image Representation Learning (SSL)~\cite{10.1109/ICCV.2015.167} and Vision Language Models (VLMs)~\cite{10.5555/3454287.3454289} can mitigate some of these issues by learning transferable representations of images via large-scale pretraining. When SSL or VLMs are deployed as pretrained backbones for a specialized classification model, the network no longer has to learn as many fundamental visual concepts from scratch. This accelerates overall training times and reduces reliance on large annotated datasets. However, fine-tuning a single backbone often locks the model into task-specific goals, hindering cross-domain generalization.

\begin{figure*}[htbp]
    \centering
    \makebox[\textwidth][c]{%
        \includegraphics[width=1\textwidth]{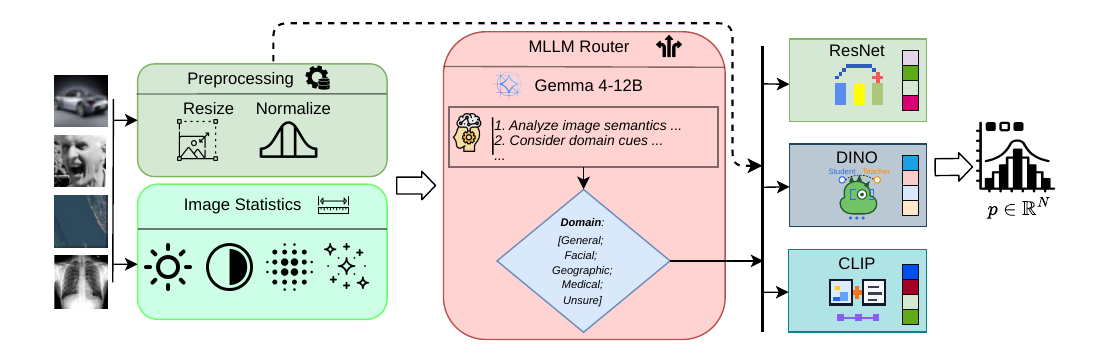}%
    }
    \caption{Overview of the ARMDIL pipeline for cross-domain image classification. Each input image is preprocessed via resizing and normalization. Additionally, an image quality assessment (blur, brightness, contrast, and noise) is computed. The raw image and quality statistics are passed jointly to an MLLM router (Gemma-4-12B), which reasons over the visual content and quality signals to assign the image to one of five domain aliases (GENERAL, FACIAL, GEOGRAPHIC, MEDICAL, or UNSURE) and routes it to the corresponding domain expert. Expert backbones (ResNet, DINO, and CLIP) are trained offline with domain-skewed sampling. Images assigned to UNSURE are redirected to the model with the best overall validation accuracy. All experts have a unified $N=38$ classification head, ensuring valid predictions regardless of routing outcome.}
    \label{fig:system_diagram}
\end{figure*}

A common strategy to overcome this single-model limitation is to deploy an ensemble of diverse architectures. However, traditional dense ensembles fail to differentiate between expert specializations, and dynamic routing methods~\cite{aljundi2017expertgate, Cruz2018DynamicClassifierSelection} are rigid and uninterpretable. Recent advancements in natural language processing offer a promising alternative: utilizing Multi-Modal Large Language Models (MLLMs) as intelligent agents \cite{osti_10451467}. Rather than exhaustively executing all models, an MLLM agent can intelligently reason about the visual input to determine which specialized backbone is the best fit for a specific instance. This effectively removes the ensemble black box, providing developers with transparent, interpretable control over how each image is processed.

We propose a framework that performs cross-dataset image classification through the lens of agentic routing (Figure \ref{fig:system_diagram}). Our specialized ``toolkit" integrates heterogeneous architectural families, including ResNets, SSL, and VLMs. Each model is trained separately on a unified label space spanning multiple distinct datasets. At inference time, the MLLM agent dynamically analyzes the input image and selects the individual backbone. By replacing black-box, training-heavy routing with MLLM prompting, our approach delivers a highly adaptable ensemble that matches specialized routers in performance while providing interpretable natural language reasoning and control.

\section{Related Work}
\label{sec:related_work}

\subsection{Vision Backbones for Image Classification}
\label{rw_backbone_classification}

Modern image classification relies on two distinct architectural paradigms. CNNs, which were popularized with AlexNet~\cite{10.5555/2999134.2999257} and expanded upon with ResNets~\cite{DBLP:journals/corr/HeZRS15, 10.5555/3042817.3043083}, leverage convolutions and pooling. ViTs~\cite{dosovitskiy2021an} replace convolutions with attention-based transformer mechanisms. Although they can achieve state-of-the-art results for specific tasks, ViTs typically require extensive data to reach optimal performance.

To alleviate dependency on meticulously labeled datasets, SSL leverages unlabeled data to learn task-agnostic representations, encoding the structural properties inherent in images. This space has evolved rapidly, from contrastive frameworks like SimCLR~\cite{10.5555/3524938.3525087} and MoCO~\cite{9157636} to non-contrastive self-distilling models like BYOL ~\cite{10.5555/3495724.3497510}, SimSiam~\cite{9578004}, Barlow Twins~\cite{zbontar2021barlow}, and DINO~\cite{9709990, oquab2024dinov, simeoni2025dinov3}. These models are often based on ResNets and/or ViTs. They can all be implemented as backbones with a trained classification head. 

VLMs, on the other hand, build a shared multimodal space between images and their natural language descriptions. CLIP~\cite{Radford2021LearningTV} and SigLIP~\cite{10377550} align image-text pairs to bridge visual patterns with semantic understanding. VLMs were originally designed for zero‑shot classification; each image is assigned the label whose text embedding has the highest cosine similarity with the image embedding. However, they have also been used as frozen backbones for supervised classification~\cite{10445007}.

\subsection{Ensembles and Cross-Dataset Robustness}
\label{rw_ensembles}

Several studies have explored ensembles that combine ViTs and/or ResNets for specific tasks~\cite{lungCancerEnsemble, Shanmugam2025, Rahul2025, AlHejri2025, DN2025, Narasimharaju2024, RemoteSensingEnsemble2026}. Because heterogeneous architectures focus on different semantic features~\cite{bernardino2023comparing, vishniakov2024convnet}, combining them can lead to more accurate and robust results. Standard fusion strategies typically concatenate feature representations or employ majority voting. More advanced approaches, such as the Neural Logit Controller (NLC)~\cite{DBLP:journals/corr/abs-2405-17139}, learn input-dependent temperatures to dynamically combine logits from multiple backbones to form a final prediction.

To avoid executing every model in an ensemble, other methods use learned routers as meta-classifiers that dynamically map inputs to individual experts~\cite{Cruz2018DynamicClassifierSelection}. Expert Gate, for instance, trains autoencoders alongside each expert to learn domain-specific representations. At test time, the input is exclusively routed to the expert whose autoencoder produces the lowest reconstruction error~\cite{aljundi2017expertgate}. By selecting a single backbone per input, these methods successfully circumvent the exhaustive computation required by standard fusion ensembles.

Despite their respective advantages, both fusion and dynamic routing methodologies suffer from fundamental structural rigidity. Adapting these systems to new tasks or domains is highly cumbersome; introducing even a new expert or domain necessitates retraining the entire NLC fusion controller or router. Furthermore, both paradigms operate as a complete black box, offering developers no interpretable or semantic control over the decision process.

Beyond these structural limitations, standard ensembles are almost exclusively designed for narrow domains. Limited work has addressed the challenge of multi-domain image classification, where a system must handle completely disjoint distributions and tasks (e.g., medical, aerial, facial, and natural images). Furthermore, to the best of our knowledge, no existing work tackles this cross-dataset heterogeneity by integrating all major deep learning architectural families (ResNets, SSL, and VLMs) into a unified ensemble.

\subsection{Multi-Modal Large Language Models}
\label{rw_mllm}

Recent advancements in natural language processing have evolved Large Language Models (LLMs) from passive text generators into goal-oriented agents. Through advanced reasoning frameworks~\cite{Guo_2025,Wang2022SelfConsistencyIC,10.5555/3600270.3602070} and the integration of external tools~\cite{10.5555/3780338.3780932, Wang2023PlanandSolvePI, osti_10451467}, modern LLMs can tackle complex tasks by assessing a prompt and dynamically routing it to the most appropriate external software module. This improves results when users need AI to perform tasks that exceed the limitations of its static training data, such as querying real-time databases, executing complex mathematical calculations, or interacting with external APIs. 

Systems, such as HuggingGPT~\cite{shen2023hugginggpt}, VisProg~\cite{gupta2023visualprogramming}, ViperGPT~\cite{suris2023vipergpt}, PhenoAssistant~\cite{chen2026phenoassistant}, and the few-shot planner developed by Meng et al.~\cite{meng2023few}, have demonstrated the viability of using LLMs as central controllers that delegate sub-tasks to specialized vision models. However, these frameworks primarily route text-based instructions to distinct functional tools (e.g., selecting an object detector versus an OCR module). Jiang et al. embed agents directly into classification pipelines \cite{jiang2025enhancing} but focus on generating human-readable concepts rather than improving accuracy with a heterogeneous ensemble.

MLLMs~\cite{liu2023visual, yang2023mmreact} extend LLMs by natively integrating visual comprehension. However, recent studies highlight that relying on MLLMs as standalone image classifiers often yields sub-optimal accuracy compared to specialized backbones~\cite{lv2025unlabeled}. The specific extension of leveraging MLLMs to intelligently route visual inputs across heterogeneous foundational architectures for cross-domain classification remains a largely underexplored gap in the literature.

\section{Methodology}
\label{sec:methodology}

We present ARMDIL: an Adaptive Router for Multi-Domain Image Classification with LLMs (Figure \ref{fig:system_diagram}). Given a set of classification models (experts), each excelling in their respective domain, ARMDIL leverages an MLLM to predict the visual domain of a given image. Our framework then routes the image to the optimal expert for classification. This approach provides an interpretable and adaptable alternative to standard ensembles, yielding natural language reasoning traces and allowing us to incorporate new knowledge without retraining.

\subsection{Cross-Domain Datasets}
\label{methodology_datasets}


To establish target domains for ARMDIL and evaluate cross-domain robustness, we first define a unified dataset that aggregates samples and labels over four public image classification benchmarks:

\begin{itemize}
    \item CIFAR10~\cite{Krizhevsky2009}: Comprising 60,000 images of diverse natural subjects (e.g., animals and vehicles), this represents a standard object-centric domain where traditional architectures like ResNets typically excel.
    \item FER2013~\cite{goodfellow2013challenges}: Containing 35,887 facial images categorized into seven emotional states, this dataset introduces a highly nuanced and difficult classification task. We expect this domain to benefit from the rich semantic understanding of VLMs or SSL backbones.
    \item EuroSAT~\cite{helber2019eurosat}: Consisting of 27,000 satellite images across 10 land-cover classes (e.g., forests, highways, and industrial buildings), this dataset tests robustness to aerial perspectives. In this domain, SSL backbones frequently demonstrate strong transferability.
    \item OrganAMNIST~\cite{yang2023medmnist, yang2021medmnist}: Featuring 58,830 medical scans categorized into 11 distinct human organs, this dataset represents a specialized domain. Medical images are conventionally addressed using CNNs, ViTs, and/or SSL. We denote this dataset as O-MNIST.
\end{itemize}

These datasets were deliberately selected to span distinct visual domains, ensuring a rigorous test of our agent's routing capabilities. To ensure stable training, our selected datasets are all similar in size, resolution, and number of classes. However, they differ significantly in domain and difficulty, requiring the use of an effective router.

\subsection{Heterogeneous Experts}
\label{methodology_training}

Candidates for our ensemble are chosen from the best-performing ResNet and ViTs (SSL and VLMs). All models use pretrained weights and undergo fine-tuning over the unified dataset. While ARMDIL can dynamically route to any classifier, we adopt this joint formulation specifically for our evaluation. The shared label space ensures that if the MLLM routes an image to a suboptimal expert, the selected classifier can still output a valid prediction.

\begin{table}[htbp]
    \caption{Data distribution for each training scenario. The first four rows denote domain-skewed (biased) configurations. For instance, the second row illustrates the distribution when the training set is heavily biased (70\%) toward CIFAR10. The last row represents the balanced baseline across all four domains.}
    \label{tab:distributions}
    \begin{tabular}{|c|cccc|}
    \hline
    \textbf{Bias} & CIFAR10 & FER2013 & EuroSAT & O-MNIST \\
    \hline
    \textbf{CIFAR10} & 70\% & 10\% & 10\% & 10\% \\
    \textbf{FER2013} & 10\% & 70\% & 10\% & 10\% \\
    \textbf{EuroSAT} & 10\% & 10\% & 70\% & 10\% \\
    \textbf{O-MNIST} & 10\% & 10\% & 10\% & 70\% \\
    \textbf{Balanced} & 25\% & 25\% & 25\% & 25\% \\
    \hline
    \end{tabular}
\end{table}

While all candidate networks are trained across the unified space, they are specialized into domain experts via weighted sampling. By configuring the data loader to guarantee a 70\% representation for a target dataset (Table \ref{tab:distributions}), we force the network to heavily bias its learned features toward that majority domain. This yields an expert highly attuned to its target distribution but intentionally sub-optimal on the underrepresented data. We also train models on a balanced uniform distribution as a baseline.

\subsubsection{ResNets}
\label{methodology_resnet}

ResNets \cite{DBLP:journals/corr/HeZRS15} excel at extracting precise local features and are highly parameter-efficient compared to ViTs and VLMs. This makes them particularly effective for domains like medical imagery, which rely heavily on distinct textures, shapes, and edges. However, their tendency to overfit to specific training distributions often degrades their performance when modeling multiple heterogeneous domains simultaneously. For our candidate experts, we initialize ResNet-50 networks with ImageNet pre-trained weights \cite{Iakubovskii:2019} and fine-tune them under the sampling regimes outlined in Table \ref{tab:distributions}.

\subsubsection{SSL}
\label{methodology_ssl}

Our second set of domain experts employs SSL-based ViT models. These models are particularly well-suited for complex visual domains, such as facial recognition, where acquiring densely labeled data is challenging but high-resolution structural understanding is still required. We utilize DINO models due to their state-of-the-art performance; their self-distillation objective produces highly robust representations that require minimal downstream adaptation. Our implementation leverages DINOv2 (ViT-L/14)~\cite{oquab2024dinov} and the recently introduced DINOv3 (ViT-L/16)~\cite{simeoni2025dinov3}. These models are pre-trained on variations of Meta's LVD datasets.

\subsubsection{VLMs}
\label{methodology_vlm}

Our final set of domain experts employs VLM-based classifiers. VLMs are well-suited for broad visual domains, where visual characteristics correspond closely to human-interpretable concepts. Furthermore, because their representations are not constrained to a fixed, predefined label space, they generalize robustly across diverse domains. We employ OpenCLIP's ViT-L/14, pretrained on LAION-2B~\cite{schuhmann2022laionb}. Rather than relying on cosine similarity for classification, we attach a linear classification head over the unified label space. To further adapt the model to our task, we keep the CLIP weights frozen and inject LoRA adapters~\cite{hu2022lora} into the MLP layers of the visual transformer blocks.

\subsection{ARMDIL}
\label{methodology_MLLM}

\subsubsection{MLLM Router}
\label{methodology_armdil}

After training the heterogeneous experts, those achieving the highest validation accuracy for each domain are deployed as tools. ARMDIL employs an MLLM as a meta-classifier and domain router. To ensure reproducibility, we use the Unsloth UD-Q5 quantization of Gemma-4-12B~\cite{gemma-4-12b}, a small model that can be run locally. Given both an image and image quality assessment as input, the MLLM selects one of five domains, routing the image to the corresponding expert for classification (Figure \ref{fig:system_diagram}).

\begin{figure}[!htbp]
    \begin{tcolorbox}[title=System Prompt, fonttitle=\bfseries, colback=gray!5, colframe=gray!50!black, arc=2mm, boxrule=0.5pt, left=4pt, right=4pt, top=4pt, bottom=4pt]
    You are an image domain classifier. You receive an image together with basic image-quality statistics for it. Use both to classify the image into exactly one domain: GENERAL, MEDICAL, GEOGRAPHIC, FACIAL, or UNSURE.
    
    \textbf{Definitions:}
    \begin{itemize}[leftmargin=*, noitemsep, topsep=2pt]
        \item \textbf{GENERAL:} Everyday objects, animals, or vehicles (natural-image content).
        \item \textbf{MEDICAL:} Medical scans, pathology slides, organ cross-sections, anatomical imagery.
        \item \textbf{GEOGRAPHIC:} Satellite imagery, aerial views, terrain environments.
        \item \textbf{FACIAL:} Human faces, facial expressions, and portraits.
        \item \textbf{UNSURE:} Use ONLY if the image genuinely does not fit any of the four domains above, or sits on a boundary between them. Do not pick UNSURE as a default; only when you truly cannot choose between GENERAL / MEDICAL / GEOGRAPHIC / FACIAL.
    \end{itemize}
    
    \textbf{Image-quality metric reference ranges} (auxiliary signals, not the sole basis for the decision):
    \begin{itemize}[leftmargin=*, noitemsep, topsep=2pt]
        \item \textbf{blur\_score:} 0.0 = no blur, 1.0 = very blurry
        \item \textbf{mean\_brightness:} 0.0 = black / very dark, 0.5 = mid-gray, 1.0 = white / very bright
        \item \textbf{contrast:} 0.0 = very low contrast, 0.03-0.07 = low contrast, 0.07-0.15 = moderate contrast, 0.15-0.25 = high contrast, 0.25+ = very high contrast
        \item \textbf{noise\_estimate:} 0.0 = nearly noise free, 0.001-0.01 = low noise, 0.05 = visibly noisy, 0.10+ = very noisy
    \end{itemize}
    
    Output ONLY the domain name, nothing else.
    \end{tcolorbox}
    \caption{MLLM prompt for ARMDIL's domain router. The prompt briefly describes the relevant image domains. It provides the model with an option to be unsure (and later apply the model with the best overall accuracy). It encourages the MLLM to think about the blur and other image details to improve its domain classification. Finally, it asks the LLM to output just the domain name at the end.}
    \label{fig:prompt}
\end{figure}

The MLLM's objective and instructions are defined in the system prompt (Figure \ref{fig:prompt}). Each dataset is assigned an alias intended to let the MLLM operate at an abstracted level. CIFAR10 is referred to as ``GENERAL", and described in terms of generic elements such as animals, vehicles, or other ``everyday objects". OrganAMNIST is assigned ``MEDICAL", described as medical scans. FER2013 is ``FACIAL", described as human faces, expressions, or portraits. Lastly, EuroSAT is assigned ``GEOGRAPHIC", described as satellite imagery, aerial views, or terrain environments. The special domain, ``UNSURE", is provided as a scapegoat for the MLLM in extreme circumstances, which will route to a non-specialized model.

To guide the MLLM's decision-making process before it outputs the final domain prediction, we leverage chain-of-thought (CoT) reasoning~\cite{10.5555/3600270.3602070}. This improves both the overall classification accuracy and interpretability of each domain prediction.

\begin{figure}[htbp]
    \centering
    \begin{minipage}[c]{0.32\linewidth}
        \centering
        \includegraphics[width=0.7\linewidth]{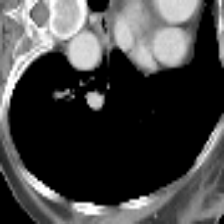}
    \end{minipage}
    \hfill
    \begin{minipage}[c]{0.65\linewidth}
        \centering
        \begin{tcolorbox}[width=\linewidth]
        \textbf{Image-quality statistics:}\\
        - blur\_score: 0.55\\
        - mean\_brightness: 0.27\\
        - contrast: 0.31\\
        - noise\_estimate: 0.0017
        \end{tcolorbox}
    \end{minipage}
    \caption{Example image quality assessment for an OrganAMNIST sample. We compute blur, brightness, contrast, and noise. These statistics are then incorporated in the MLLM prompt (Figure \ref{fig:prompt})}
    \label{fig:example_organ}
\end{figure}

Furthermore, to supply the MLLM with additional context for its reasoning, we provide an image quality assessment (Figure \ref{fig:example_organ}) in each prompt. This assessment includes explicit metrics for image blur, brightness, contrast, and noise. The blur score is calculated using a no-reference perceptual blur metric~\cite{creteroffet:hal-00232709}. Brightness and contrast are calculated as the mean and standard deviation of the image cast to grayscale, respectively. Lastly, noise is estimated based on the median absolute deviation of the wavelet detail coefficients~\cite{10.1093/biomet/81.3.425}. These low-level visual characteristics serve as valuable heuristic cues for domain differentiation; for instance, a medical scan often exhibits a distinct blur profile and inherently lower brightness than a standard photograph of an animal or human face.

\subsubsection{Ensemble Baselines}
\label{methodology_baseline}
Two ensemble baselines are evaluated alongside ARMDIL to assess the benefit of MLLM-based routing. The first is a simple Majority Vote (MV) ensemble, where each backbone model independently predicts a class label for a given input and the final output is determined by a plurality vote. While this is a standard training-free approach, it suffers from many limitations. First, since it naively weights all experts equally, a majority vote ensemble fails to leverage the domain-specific strengths of individual models. Experts that perform poorly outside their specialized domain may develop shared biases, causing them to confidently converge on the same incorrect prediction. Second, this consensus mechanism breaks down entirely if any of the experts are trained on restricted or disjoint label spaces, significantly limiting future adaptability.

The second baseline substitutes ARMDIL’s MLLM with a neural network router (NN-Router). This explicitly classifies the visual domain of each input to route it to the appropriate domain-specific expert. While this learned routing function achieves competitive accuracy on its training distribution, it also suffers from significant limitations compared to our approach. First, it operates as a black box, lacking the language-driven interpretability inherent to an MLLM. Second, because the domain classifier must be explicitly trained on a predefined set of classes, it is fundamentally rigid; introducing or modifying input domains necessitates costly retraining or fine-tuning.

\section{Results}
\label{sec:results}

We evaluate ARMDIL to establish that its dynamic routing performs comparably to traditional baselines while offering clear advantages in adaptability and interpretability. Throughout this analysis, performance is measured using overall and per-dataset top-1 accuracy and F1 Scores. We first characterize the unique strengths and weaknesses of the heterogeneous backbones by evaluating them individually. Leveraging the strongest classification models for each domain, we then benchmark ARMDIL against traditional ensemble mechanisms. Finally, we conduct ablation studies to isolate the impact of the router's core components.

\begin{table}[!hbtp]
\footnotesize 
\setlength{\tabcolsep}{3pt} 
\renewcommand{\arraystretch}{1.05} 
\caption{Per-dataset cross-distribution performance across all backbone architectures. The training distribution corresponds to the domain-skewed sampling used during training (Table \ref{tab:distributions}). The test set corresponds to the dataset each trained model was evaluated on, with ``Overall'' reporting global metrics on the unified test dataset. Values represent top-1 accuracy (left) and F1 Score (right). Bold entries indicate optimal cross-distribution configuration settings within each backbone.}
\label{tab:all_heterogeneous_backbones}

\begin{tabular}{ll ccccc}
\toprule
& & \multicolumn{5}{c}{\textbf{Training Distribution (Top-1 Acc \% / F1 Score \%)}} \\
\cmidrule(lr){3-7}
\textbf{Backbone} & \textbf{Test Set} & \textbf{CIFAR10} & \textbf{EuroSAT} & \textbf{FER2013} & \textbf{O-MNIST} & \textbf{Balanced} \\
\midrule

\multirow{5}{*}{\textbf{ResNet-50}} 
  & CIFAR10  & \textbf{94.36} / \textbf{94.36} & 91.54 / 91.53 & 88.17 / 88.08 & 90.05 / 90.19 & 92.90 / 92.94\\
  & EuroSAT  & 96.07 / 96.01 & \textbf{98.17} / \textbf{98.14} & 95.96 / 96.00 & 94.98 / 94.92 & 97.17 / 97.18\\
  & FER2013  & 60.32 / 55.18 & 64.61 / 63.31 & \textbf{66.40} / \textbf{65.57} & 58.14 / 53.06 & 61.86 / 58.79\\
  & O-MNIST  & 96.01 / 95.78 & 95.89 / 95.58 & 94.48 / 94.11 & \textbf{97.20} / \textbf{96.97} & 95.15 / 94.79\\
  & \textbf{Overall}  & 86.04 / 85.48 & 86.56 / 86.76 & 85.55 / 85.68 & 84.87 / 84.86 & \textbf{86.69} / \textbf{86.81}\\

\midrule
\multirow{5}{*}{\makecell[l]{\textbf{DINOv2}\\\textbf{(ViT-L/14)}}} 
  & CIFAR10  & 99.24 / 99.27 & 99.06 / 99.17 & 99.13 / 99.22 & 99.04 / 99.14 & \textbf{99.24} / \textbf{99.28} \\
  & EuroSAT  & 96.02 / 95.96 & \textbf{96.65} / \textbf{96.55} & 95.91 / 95.85 & 95.74 / 95.63 & 96.30 / 96.20 \\
  & FER2013  & 64.06 / 60.22 & 64.03 / 59.73 & 65.32 / 62.56 & 63.26 / 58.68 & \textbf{65.34} / \textbf{62.57} \\
  & O-MNIST  & 90.90 / 90.30 & 90.12 / 89.52 & 90.18 / 89.47 & \textbf{92.52} / \textbf{92.00} & 91.83 / 91.34 \\
  & \textbf{Overall}  & 87.50 / 87.48 & 87.69 / 87.63 & \textbf{87.95} / \textbf{87.96} & 87.27 / 87.15 & 87.95 / 87.94 \\

\midrule
\multirow{5}{*}{\makecell[l]{\textbf{DINOv3}\\\textbf{(ViT-L/16)}}} 
  & CIFAR10  & 99.02 / 99.02 & 98.97 / 99.02 & 99.00 / 99.03 & 99.02 / 99.06 & \textbf{99.05} / \textbf{99.06} \\
  & EuroSAT  & 96.87 / 96.81 & \textbf{97.48} / \textbf{97.42} & 96.61 / 96.57 & 96.46 / 96.40 & 97.24 / 97.19 \\
  & FER2013  & 68.93 / 64.58 & 68.97 / 65.16 & \textbf{70.49} / \textbf{67.75} & 67.99 / 63.71 & 69.82 / 66.63 \\
  & O-MNIST  & 91.70 / 91.26 & 91.44 / 91.02 & 91.10 / 90.52 & \textbf{93.29} / \textbf{93.17} & 92.52 / 92.22 \\
  & \textbf{Overall}  & 88.90 / 88.77 & 89.16 / 89.05 & 88.90 / 88.83 & 89.16 / 89.05 & \textbf{89.61} / \textbf{89.52} \\

\midrule
\multirow{5}{*}{\makecell[l]{\textbf{CLIP}\\\textbf{(ViT-L/16)}}} 
  & CIFAR10  & \textbf{98.33} / \textbf{98.33} & 97.89 / 97.89 & 97.20 / 97.20 & 97.48 / 97.48 & 97.91 / 97.91 \\
  & EuroSAT  & \textbf{98.56} / \textbf{98.51} & 98.52 / 98.46 & 97.85 / 97.78 & 98.30 / 98.27 & 98.11 / 98.04 \\
  & FER2013  & 64.43 / 52.03 & 64.36 / 51.87 & 67.21 / \textbf{62.42} & 65.20 / 52.33 & \textbf{67.58} / 56.77 \\
  & O-MNIST  & 90.31 / 82.99 & 90.72 / 83.33 & 84.95 / 77.92 & \textbf{96.59} / \textbf{96.34} & 96.54 / 96.08 \\
  & \textbf{Overall}  & 84.12 / 82.19 & 83.77 / 81.75 & 83.54 / 82.63 & 86.49 / 85.21 & \textbf{87.53} / \textbf{86.58} \\
\bottomrule
\end{tabular}
\end{table}

\subsection{Top Classifier Performance}
\label{results_overall}

To shed light on the strengths and weaknesses of each model, we display the results of all the trained experts (Table \ref{tab:all_heterogeneous_backbones}) for each architectural family and training distribution. 

Our ResNet classifiers demonstrated high domain-specific accuracy, achieving strong results on EuroSAT and OrganAMNIST. However, they yielded slightly lower overall performance across the combined datasets and exhibited sensitivity to shifts in the training distribution. This variance indicates that while highly effective for specific tasks, these traditional convolutional architectures face limitations in learning a universally shared representation across highly disparate visual domains.

In comparison, the DINO architectures provided a modest improvement in broad generalization. Both DINO variants achieved near-perfect accuracy on CIFAR10 and showed greater resilience when the training distribution was altered. This improvement was particularly significant with DINOv3, which outperformed all other architectures by 2.91\% on FER2013, the most challenging dataset. These findings suggest that the attention-based backbones and self-supervised pre-training yield feature representations better equipped to handle complex visual domains without sacrificing overall stability.

Surprisingly, the CLIP models outperformed all other architectures on the EuroSAT dataset. We hypothesize that this is because aerial imagery classification relies more heavily on scene-level semantics than on local textures and shapes. Similar to the DINO models, CLIP also demonstrated robust resilience to distribution shifts and significantly outperformed the ResNet baseline on CIFAR10.

Table \ref{results:top_models} displays the best-performing classifiers for each dataset across all trained experts. These models will serve as domain experts for ARMDIL. Ultimately, each expert excels on specific tasks but does not generalize as well across all domains, aligning with the standard premise of ensemble learning.

\begin{table}[h]
    \caption{The best-performing models for each dataset. ``Overall" reports the model that performed best on the unified test set. These are the experts that ARMDIL and our baselines employ.}
    \label{results:top_models}
    \begin{tabular}{|c|cccc|}
    \hline
     \textbf{Dataset} & \textbf{Best Model} & \textbf{Bias} & \textbf{Acc} & \textbf{Macro F1}\\
    \hline
     CIFAR10  & DINOv2 & Balanced   & 99.24 & 99.28\\
     EuroSAT  & CLIP   & CIFAR10    & 98.56 & 98.51\\
     FER2013  & DINOv3 & FER2013    & 70.49 & 67.75\\
     O-MNIST  & ResNet-50    & O-MNIST & 97.20 & 96.97\\
     \textbf{Overall}  & DINOv3       & Balanced   & 89.61 & 89.52\\
        \hline
    \end{tabular}
\end{table}

\subsection{ARMDIL}
\label{results_armdil}

\subsubsection{Domain Routing}
To evaluate ARMDIL, we first compare the MLLM's routing predictions against the ground-truth domains, with the resulting confusion matrix displayed in Table \ref{tab:domain_classification}. The MLLM achieved exceptional accuracy on FER2013, yielding a true positive rate of $99.82\%$. However, performance on EuroSAT reflected less confidence, with $12.72\%$ of the images routed to the UNSURE domain. This uncertainty likely stems from the blurry and abstract nature of the aerial imagery, which often lacks distinctive visual features. 

\begin{table}[htb]
    \centering 
    \caption{Domain routing accuracy of ARMDIL's MLLM}
    \label{tab:domain_classification}
    \setlength{\tabcolsep}{4pt}
    \begin{tabular}{lccccc}
        \toprule
        & \multicolumn{5}{c}{\textbf{Predicted Domain}} \\
        \cmidrule(lr){2-6} 
        \textbf{True Domain} & \textbf{GEN.} & \textbf{GEO.} & \textbf{FAC.} & \textbf{MED.} & \textbf{UNSURE} \\
        \midrule
        CIFAR-10 & \textbf{97.80} & 0.43 & 0.61 & 0.93 & 0.23 \\
        EuroSAT  & 5.24 & \textbf{78.20} & 0.06 & 3.78 & 12.72 \\
        FER2013  & 0.10 & 0.00 & \textbf{99.82} & 0.01 & 0.07 \\
        O-MNIST  & 0.69 & 0.78 & 1.55 & \textbf{95.21} & 1.76 \\
        \bottomrule
    \end{tabular}
\end{table}

Overall, these zero-shot results are highly promising, as the MLLM router operated entirely via prompt-based inference without requiring domain-specific training. Refining the descriptions in the prompt or fine-tuning the MLLM, will likely further improve overall routing accuracy.

\subsubsection{Final Classification}

\begin{table}[h]
\caption{Per-dataset performance of the best individual backbone (DINOv3, unbiased), ARMDIL, the two ensemble baselines, and the theoretical oracle router. Each cell reports accuracy (top) and per-dataset macro F1 over that test set's label set. \textbf{Overall} reports global accuracy on the unified dataset and the mean of per-dataset macro F1 values across unified-dataset domains.}
\label{tab:ensemble_results}
\setlength{\tabcolsep}{2.5pt}
\renewcommand{\arraystretch}{1.15}
    \begin{tabular}{lcccc|c}
        \toprule
        & \multicolumn{5}{c}{\textbf{Method}} \\
        \cmidrule(lr){2-6}
        \textbf{Dataset} & \textbf{ARMDIL} & DINOv3 & MV & NN-Router & \textit{Oracle} \\
        \midrule
        
        CIFAR10
        & \makecell{99.02\\99.05}
        & \makecell{99.05 \\ 99.06}
        & \makecell{98.87 \\ 98.87}
        & \makecell{\textbf{99.13} \\ \textbf{99.18}}
        & \makecell{\textit{99.24}\\\textit{99.28}} \\
        
        \midrule
        
        EuroSAT
        & \makecell{97.91\\97.88}
        & \makecell{97.24 \\ 97.19}
        & \makecell{98.42 \\ 98.38}
        & \makecell{\textbf{98.56} \\ \textbf{98.52}}
        & \makecell{\textit{98.56}\\\textit{98.51}} \\
        
        \midrule
        
        FER2013
        & \makecell{\textbf{70.49}\\\textbf{67.73}}
        & \makecell{69.82 \\ 66.63}
        & \makecell{68.35 \\ 64.64}
        & \makecell{69.81 \\ 66.98}
        & \makecell{\textit{70.49}\\\textit{67.75}} \\
        
        \midrule
        
        O-MNIST
        & \makecell{96.69\\96.38}
        & \makecell{92.52 \\ 92.22}
        & \makecell{96.25 \\ 95.84}
        & \makecell{\textbf{97.22} \\ \textbf{96.95}}
        & \makecell{\textit{97.20}\\\textit{96.97}} \\
        
        \midrule
        
        \textbf{Overall}
        & \makecell{90.78\\\textbf{90.71}}
        & \makecell{89.61 \\ 89.52}
        & \makecell{90.47 \\ 89.43}
        & \makecell{\textbf{91.18} \\ 90.41}
        & \makecell{\textit{91.04}\\\textit{90.99}} \\
        
        \bottomrule
    \end{tabular}
\end{table}

Table \ref{tab:ensemble_results} reports the per-dataset and overall results for ARMDIL (Section \ref{methodology_armdil}), the best overall expert (DINOv3), and two standard ensemble baselines (Section \ref{methodology_baseline}). To establish a performance ceiling, we also include a theoretical ``oracle" router, which mirrors ARMDIL's configuration but assumes perfect routing accuracy.

ARMDIL achieves an impressive 90.78\% accuracy on the unified test set, outperforming the strongest individual backbone (the balanced DINOv3 expert) by 1.17\%. It demonstrates significant gains over DINOv3 on OrganAMNIST and FER2013, while marginally underperforming on CIFAR10, a dataset where the balanced DINOv3 model already achieves parity with the specialized CIFAR10 expert.

ARMDIL surpasses the majority vote ensemble by $0.31\%$ in overall accuracy and $1.28\%$ in macro F1 Score. It outperforms this baseline across all datasets with the exception of EuroSAT, a drop that directly aligns with the MLLM router's lowest confidence scores. This limitation could likely be mitigated by refining the prompt to include a more comprehensive description of the ``GEOGRAPHIC" domain and explicitly instructing the model to minimize ``UNSURE" predictions.

Notably, ARMDIL outperforms all ensemble baselines by $0.68\%$ on FER2013, which represents the most challenging dataset in our evaluation. This strong performance suggests that ARMDIL's architectural flexibility will yield even greater advantages over standard ensembles in highly complex domains where specialized expertise is indispensable.

Most importantly, ARMDIL proves highly competitive even against a fully trained routing network. Compared to the near-optimal NN-Router, ARMDIL actually improves the overall F1 Score by $0.30\%$ while trailing in overall accuracy by a mere $0.40\%$. Crucially, ARMDIL achieves this baseline without the costly additional training overhead required by an NN-Router.

\subsection{Adaptability and Interpretability}
\label{sec:interpretability}

\begin{figure}[htbp]
    \centering
    \includegraphics[width=0.3\linewidth]{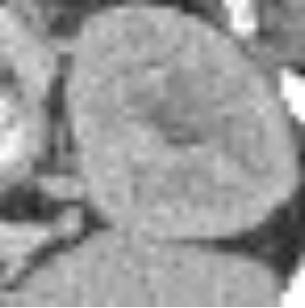}
    
    \begin{tcolorbox}[width=\linewidth, colback=gray!5, colframe=gray!50!black, arc=2mm, boxrule=0.5pt, left=3pt, right=3pt, top=3pt, bottom=3pt]
    The user wants me to classify an image into one of five categories: GENERAL, MEDICAL, GEOGRAPHIC, FACIAL, or UNSURE.
    
    \begin{enumerate}[leftmargin=*, noitemsep, topsep=2pt, partopsep=0pt]
        \item \textbf{Analyze the image:}
        \begin{itemize}[leftmargin=*, noitemsep, topsep=2pt]
            \item The image shows a grayscale, cross-sectional view of what appears to be an internal body part.
            \item It looks like a CT scan or MRI of a lung or chest area.
            \item There are distinct anatomical structures, including what looks like lung tissue (darker areas) and soft tissue/vessels (lighter areas).
            \item This fits the definition of MEDICAL imagery (scans, pathology, etc.).
        \end{itemize}
        
        \item \textbf{Analyze the statistics:}
        \begin{itemize}[leftmargin=*, noitemsep, topsep=2pt]
            \item blur\_score: 0.53 (somewhat blurry, common in some medical scans).
            \item mean\_brightness: 0.38 (mid-dark).
            \item contrast: 0.20 (high contrast, typical for medical imaging to distinguish tissue).
            \item noise\_estimate: 0.0011 (low noise).
        \end{itemize}
        
        \item \textbf{Compare with definitions:}
        \begin{itemize}[leftmargin=*, noitemsep, topsep=2pt]
            \item GENERAL: Not everyday objects.
            \item MEDICAL: Matches (CT/MRI scan).
            \item GEOGRAPHIC: Not satellite/aerial.
            \item FACIAL: Not a human face.
            \item UNSURE: The image is clearly a medical scan.
        \end{itemize}
        
        \item \textbf{Final Decision:} MEDICAL.
    \end{enumerate}
    \end{tcolorbox}
    
    \caption{Example reasoning trace of ARMDIL's MLLM. This is for a randomly selected OrganAMNIST image of a left kidney.}
    \label{fig:reasoning_trace}
\end{figure}

A primary advantage of ARMDIL lies in its inherent adaptability. Integrating new domains, expert models, or prior knowledge requires only a straightforward modification to the routing prompt (Figure \ref{fig:prompt}). For example, if operational requirements expand to include military vehicles, the user simply appends a ``VEHICLES'' domain to the prompt, provides a brief semantic definition, and assigns a corresponding expert classifier. In contrast, traditional ensemble architectures generally require a significant overhaul when introducing new categories, often necessitating the retraining of the entire routing module.

Furthermore, ARMDIL's routing mechanism offers enhanced interpretability by generating explicit, natural-language explanations for its decisions. For instance, as demonstrated in Figure \ref{fig:reasoning_trace}, when presented with an OrganAMNIST kidney image, the MLLM generates a step-by-step reasoning trace. It visually identifies the input as a medical scan, albeit misidentifying the kidney as a lung, and verifies this assessment using auxiliary image statistics before finalizing its domain selection. While the MLLM lacks fine-grained anatomical precision, its high-level reasoning remains robust enough to correctly route the image to the MEDICAL domain, where a specialized expert can accurately perform the downstream classification.

\subsection{Alternative Routing Configurations}
\label{sec:ablation}

To evaluate the efficacy of different MLLM routing strategies, we analyze alternative configurations. Specifically, we investigate the tradeoffs between chain-of-thought reasoning~\cite{10.5555/3600270.3602070}, self-consistency~\cite{Wang2022SelfConsistencyIC}, and the image-quality statistics (Figure \ref{fig:example_organ}).

To evaluate self-consistency as an alternative routing mechanism, we aggregate domain classifications from four independent MLLM prompts via majority voting. To accommodate the computational overhead of deploying the Gemma-4-12B model across multiple parallel runs, we omit CoT reasoning for these specific evaluations. This allows us to explicitly contrast the benefits of multi-pass consensus against single-pass logical deduction.

\begin{table}[h]
    \caption{Domain classification accuracy (\%) for the evaluated LLM routing ablation configurations. The tested ensembles include the original ARMDIL model, ARMDIL with self-consistency and no CoT reasoning, and ARMDIL with self-consistency and no CoT reasoning or image-quality statistics.}
    \label{tab:ablation_routing}
    \setlength{\tabcolsep}{6pt}
    \renewcommand{\arraystretch}{1}
    \begin{tabular}{lccc}
        \toprule
        \textbf{Dataset} & \textbf{\shortstack{\textbf{ARMDIL}}} & \textbf{\shortstack{SC\\w/o CoT}} & \textbf{\shortstack{SC\\ w/o CoT \& IQ}} \\
        \midrule
        CIFAR10 & \textbf{97.80} & 97.78 & 97.06 \\
        EuroSAT  & 78.20 & 82.17 & \textbf{83.87} \\
        FER2013  & \textbf{99.82} & 99.80 & \textbf{99.82} \\
        O-MNIST  & \textbf{95.21} & 94.88 & 94.72 \\
        \textbf{Average} & 92.76 & 93.66 & \textbf{93.87} \\
        \bottomrule
    \end{tabular}
\end{table}

The routing accuracy for each configuration is detailed in Table \ref{tab:ablation_routing}. While self-consistency slightly reduces accuracy on CIFAR10, FER2013, and OrganAMNIST, with the image-quality statistics often compounding these marginal declines, it provides a clear performance boost on EuroSAT. Specifically, self-consistency increases EuroSAT's routing accuracy by $3.97\%$, with an additional $1.70\%$ gain realized when image-quality statistics are omitted. Given that EuroSAT is the dataset the MLLM is most uncertain about, self-consistency successfully resolves this hesitation by filtering out sporadic ``UNSURE" predictions. Additionally, omitting the image-quality statistics is advantageous for this specific domain because standard quality metrics often mischaracterize the uniform textures and lack of prominent edges in aerial imagery as blur.

\begin{table}[h]
    \caption{Per-dataset performance of the ensemble using three Gemma-4-12B LLM routing configurations. Each cell reports classification accuracy (left) and per-dataset macro F1 (right). Overall reports global accuracy on the unified dataset and the mean of per-dataset macro F1 values across unified-dataset domains.}
    \label{tab:llm_router_ablation}
    \setlength{\tabcolsep}{5pt}
    \renewcommand{\arraystretch}{1}
    \begin{tabular}{lccc}
        \toprule
        \textbf{Dataset}
        & \makecell{\textbf{ARMDIL}}
        & \makecell{\textbf{SC}\\\textbf{w/o CoT}}
        & \makecell{\textbf{SC}\\\textbf{w/o CoT \& IQ}} \\
        \midrule
        
        CIFAR10
        & \makecell{99.02 / 99.05}
        & \makecell{\textbf{99.07} / \textbf{99.09}}
        & \makecell{99.01 / 99.03} \\
        
        EuroSAT
        & \makecell{97.91 / 97.88}
        & \makecell{98.02 / 97.98}
        & \makecell{\textbf{98.07} / \textbf{98.04}} \\
        
        FER2013
        & \makecell{70.49 / 67.73}
        & \makecell{\textbf{70.51} / \textbf{67.77}}
        & \makecell{70.49 / 67.76} \\
        
        O-MNIST
        & \makecell{\textbf{96.69} / \textbf{96.38}}
        & \makecell{96.67 / 96.21}
        & \makecell{96.59 / 96.04} \\
        
        \textbf{Overall}
        & \makecell{\textbf{90.78} / \textbf{90.71}}
        & \makecell{90.71 / 90.65}
        & \makecell{90.75 / 90.68} \\
        
        \bottomrule
    \end{tabular}
\end{table}

Table \ref{tab:llm_router_ablation} details the downstream classification accuracy and F1 Scores for each ablation. While the self-consistency models achieve higher overall routing accuracies, their classification accuracies and F1 Scores marginally lag behind ARMDIL. This discrepancy is largely due to ARMDIL's built-in fault tolerance. The only domain the MLLM struggles with is EuroSAT, and those misclassified inputs are subsequently directed to the balanced expert, which still maintains robust classification on satellite imagery.

\section{Conclusion}
\label{sec:conclusion}

In this paper, we introduced ARMDIL, an ensemble that replaces rigid meta-classifiers with a zero-shot MLLM agent for multi-domain image classification. ARMDIL effectively navigates the strengths and vulnerabilities of each expert within a heterogeneous ensemble of CNNs, SSL models, and VLMs. Our proposed model outperforms the standard majority vote ensemble and is highly competitive with a specialized neural network-based router, all without requiring any routing-specific training. Notably, ARMDIL achieves these impressive results while offering significant added benefits in adaptability and interpretability.

While ARMDIL represents a significant step forward, there remains room for improvement. Because small LLMs struggle to reason~\cite{lanham2023faithfulness}, future work will replace our local MLLM with a larger, industry-standard one. Furthermore, modifying the prompt and fine-tuning the MLLM could yield immediate performance gains. Lastly, evaluating ARMDIL on a broad set of larger and more difficult datasets will further demonstrate its capabilities, paving the way for general-purpose vision systems.

\clearpage

\begin{appendices}
\section{Implementation Details}
\subsection{Hardware}
\label{sec:hardware}
The ResNet and DINO classification models were trained across NVIDIA A100 GPUs. The OpenCLIP VLM was fine-tuned on a single NVIDIA RTX 4090 GPU. Our Gemma-4-12B MLLM was run on a local GPU with 16GB of VRAM for domain routing.

\subsection{Training Specifications and Hyperparameters}

\subsubsection{Heterogeneous Experts}
\label{sec:training_specs}
Our ResNet and DINO models maintain the same training pipeline and hyperparameter configuration (Table \ref{tab:training_config}) for each distribution. Training images are augmented using geometric and photometric transformations. Each transformation is randomly applied to the images, but with increased probability for certain domains.

\begin{table}[h]
\caption{Training configuration used across the ResNet and DINO backbone experiments.}
\label{tab:training_config}
\setlength{\tabcolsep}{5pt}
\renewcommand{\arraystretch}{1.15}
\begin{tabular}{ll}
\toprule
\textbf{Hyperparameter} & \textbf{Value} \\
Batch size & 32 \\
Max Epochs& 100 \\
$\eta_{max}$& $1 \times 10^{-4}$ \\
LR warmup epochs & 5 \\
LR decay epochs & 50 \\
Optimizer precision & Mixed precision (AMP) \\
Early stopping patience & 3 \\
Early stopping min delta & 0.0 \\
\hline
\end{tabular}
\end{table}

Models were optimized through the AdamW optimizer with default hyper-parameters and a combination of focal and weighted class-balanced cross-entropy (WCB-CE) loss.

The learning rate, $\eta$, was scheduled using a linear warm-up from zero to a maximum learning rate, $\eta_{max}$, for 5 epochs. This warm-up was followed by a cosine decay of 50 epochs to the minimum learning rate, $\eta_{min}$, and kept for the remainder of training:
\begin{equation}
\eta =
\begin{cases}
\eta_{\max}\dfrac{t}{T_w}, &
{\scriptsize \text{ } 0 \le t < T_w}, \\[0.15em]

{\scriptsize
\begin{aligned}[t]
\eta_{\min}
&+ \dfrac{1}{2}\left(1 + \cos\left(\pi\dfrac{t - T_w}{T_d}\right)\right) \\
&\times (\eta_{\max} - \eta_{\min})
\end{aligned}}
&
{\scriptsize \text{ } T_w \le t \le T_w + T_d}, \\[0.15em]

\eta_{\min}, &
{\scriptsize \text{ } t > T_w + T_d}
\end{cases}
\label{eq:lr}
\end{equation}
where, $t$ denotes the training epoch, $T_w=10$ is the warmup period, $T_d=50$ is the cosine decay period, $\eta_{\max}=10^{-4}$ is the maximum decoder learning rate, and $\eta_{\min}$ is $10\%$ of $\eta_{max}$. Models can train for a maximum of 100 epochs. However, an early stopping mechanism was implemented to end model training if there was no improvement to validation loss.

\subsubsection{NN-Router Baseline}
Our NN-Router is trained with a ResNet-18 backbone. A 4-class head is used to determine which of the four domains (General, Geo-spatial, Medical, Facial) an image belongs to. Because these domains exhibit pronounced visual differences, this lightweight architecture proved highly effective, achieving a domain classification accuracy of $99.5\%$.

\section{Additional Results}
\subsection{Heterogeneous Experts}


\begin{figure}[htbp]
    \centering
    \includegraphics[width=0.8\linewidth]{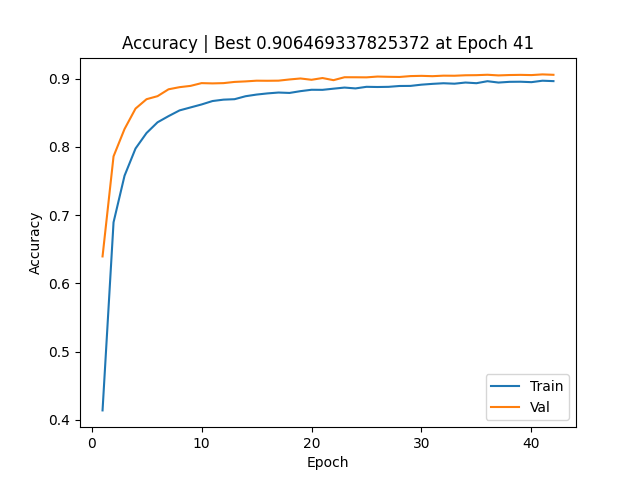}
    \caption{The training and validation accuracy of the balanced DINOv3 model at each epoch.}
    \label{fig:training_dino3l_acc}
\end{figure}


\begin{table}[h]
\caption{Per-dataset performance across training distributions of the ResNet-101 model. Each cell reports Top-1 accuracy and macro F1 Score (\%). The final column denotes performance under a balanced training distribution.}
\label{tab:resnet101_results_reformatted}
\setlength{\tabcolsep}{1.5pt}
\renewcommand{\arraystretch}{1.15}
    \begin{tabular}{lccccc}
    \toprule
    & \multicolumn{5}{c}{\textbf{Training distribution (Top-1 Acc \% / F1 Score \%)}} \\
    \cmidrule(lr){2-6}
    \textbf{Test set} & \textbf{CIFAR10} & \textbf{EuroSAT} & \textbf{FER2013} & \textbf{O-MNIST} & \textbf{Balanced} \\
    \midrule
    CIFAR10        & \makecell{\textbf{94.53}\\\textbf{94.55}} & \makecell{92.77\\92.80} & \makecell{90.91\\90.99} & \makecell{91.15\\91.24} & \makecell{92.85\\92.90} \\
    \midrule
    EuroSAT        & \makecell{96.15\\96.17} & \makecell{\textbf{98.35}\\\textbf{98.34}} & \makecell{96.17\\96.10} & \makecell{96.63\\96.64} & \makecell{97.35\\97.31} \\
    \midrule
    FER2013        & \makecell{63.54\\59.68} & \makecell{66.22\\65.17} & \makecell{65.24\\63.21} & \makecell{65.85\\63.28} & \makecell{\textbf{66.59}\\\textbf{65.28}} \\
    \midrule
    O-MNIST        & \makecell{94.54\\94.37} & \makecell{95.76\\95.60} & \makecell{92.09\\91.18} & \makecell{\textbf{97.02}\\\textbf{97.09}} & \makecell{95.87\\95.83} \\
    \midrule
    \textbf{Overall} & \makecell{86.85\\86.75} & \makecell{87.44\\87.51} & \makecell{85.10\\85.17} & \makecell{87.47\\87.44} & \makecell{\textbf{87.60}\\\textbf{87.60}} \\
    \bottomrule
    \end{tabular}
\end{table}

To better understand the performance of each expert (Section \ref{results_overall}), we analyze the loss, accuracy, and macro F1 Scores. As an example, we show the results for the balanced DINOv3 model on the unified dataset (Figure \ref{fig:training_dino3l_acc})

In addition to training the experts displayed in Table \ref{tab:all_heterogeneous_backbones}, we also trained ResNet-101 and MobileCLIP-S1 based experts. These results are shown in Tables \ref{tab:resnet101_results_reformatted} and \ref{tab:mobile_clip_results_reformatted}. Since these architectures did not perform as well as the ResNet-50 and CLIP ViT-L/14 experts on specific domains, we excluded them from our paper.

\begin{table}[h]
\caption{Per-dataset performance across training distributions of the MobileCLIP-S1 model. Each cell reports Top-1 accuracy and macro F1 Score (\%). The final column denotes performance under a balanced training distribution.}
\label{tab:mobile_clip_results_reformatted}
\setlength{\tabcolsep}{1.5pt}
\renewcommand{\arraystretch}{1.15}
    \begin{tabular}{lccccc}
    \toprule
    & \multicolumn{5}{c}{\textbf{Training distribution (Top-1 Acc \% / F1 Score \%)}} \\
    \cmidrule(lr){2-6}
    \textbf{Test set} & \textbf{CIFAR10} & \textbf{EuroSAT} & \textbf{FER2013} & \textbf{O-MNIST} & \textbf{Balanced} \\
    \midrule
    CIFAR10        & \makecell{\textbf{89.34}\\\textbf{81.18}} & \makecell{79.38\\71.89} & \makecell{70.60\\63.38} & \makecell{74.81\\67.95} & \makecell{88.49\\80.54} \\
    \midrule
    EuroSAT        & \makecell{82.83\\81.24} & \makecell{\textbf{97.70}\\\textbf{97.65}} & \makecell{93.41\\84.80} & \makecell{94.61\\94.52} & \makecell{97.11\\97.06} \\
    \midrule
    FER2013        & \makecell{26.08\\6.97} & \makecell{50.18\\31.36} & \makecell{\textbf{58.60}\\\textbf{42.83}} & \makecell{44.05\\25.64} & \makecell{55.46\\39.35} \\
    \midrule
    O-MNIST        & \makecell{75.72\\63.33} & \makecell{90.13\\79.30} & \makecell{83.11\\73.22} & \makecell{\textbf{95.26}\\\textbf{86.69}} & \makecell{93.60\\84.62} \\
    \midrule
    \textbf{Overall} & \makecell{71.22\\66.20} & \makecell{81.37\\78.11} & \makecell{77.03\\75.00} & \makecell{81.00\\77.26} & \makecell{\textbf{86.02}\\\textbf{83.82}} \\
    \bottomrule
    \end{tabular}
\end{table}

\subsection{Ablation Studies}

\begin{table}[h]
\caption{Average win ratio and unanimous rate across datasets for the ARMDIL model with self-consistency and no CoT reasoning.}
\label{tab:win_ratios}
    \begin{tabular}{lcc}
    \hline
    \textbf{Dataset} & \textbf{Winning Vote Ratio} & \textbf{\% Unanimous} \\
    \hline
    CIFAR-10 & 99.80& 99.37\\
    EuroSAT & 98.81& 96.44\\
    FER2013 & 100& 100\\
    O-MNIST & 99.25& 97.89\\
    \textbf{Average} & 99.47 & 98.43 \\
    \hline
    \end{tabular}
\end{table}

To determine the effects of self-consistency, we also investigate the variance in the ARMDIL model's domain predictions using four independent runs per image without CoT reasoning. As shown in Table \ref{tab:win_ratios}, $98.43\%$ of trials produced unanimous predictions, with $99.47\%$ of individual runs matching the final majority vote. These results indicate that, in the absence of CoT reasoning, self-consistency has no significant impact on domain predictions.

Finally, Table \ref{tab:routing_confusion_matrices} expands upon the results of Table \ref{tab:ablation_routing} by presenting the full confusion matrices for the domain predictions from our ablation studies. Notably, introducing self-consistency and removing CoT reasoning resulted in significantly fewer ``UNSURE'' predictions. While this yielded performance gains in the ``GEOGRAPHIC'' domain, it also caused a slight degradation in the ``MEDICAL'' domain.

\begin{table}[h]
\caption{MLLM routing confusion matrices for the ablation studies. Rows correspond to the true domain and columns to the predicted domain. Domain abbreviations denote CIFAR10 (GEN.), Fer2013 (FAC.), EuroSAT (GEO.), and O-MNIST (MED.).}
\label{tab:routing_confusion_matrices}
\setlength{\tabcolsep}{3pt}
\renewcommand{\arraystretch}{1.05}

\begin{tabular}{lccccc}
\toprule
& \multicolumn{5}{c}{\textbf{ARMDIL}} \\
\cmidrule(lr){2-6}
\textbf{True} / \textbf{Pred.} & GEN. & FAC. & GEO. & MED. & UNSURE \\
\midrule
GEN. & 9780 & 61 & 43 & 93 & 23 \\
FAC. & 7 & 7165 & 0 & 1 & 5 \\
GEO. & 283 & 3 & 4223 & 204 & 687 \\
MED. & 123 & 276 & 139 & 16927 & 313 \\
\bottomrule
\end{tabular}

\vspace{0.5em}

\begin{tabular}{lccccc}
\toprule
& \multicolumn{5}{c}{\textbf{SC w/o CoT}} \\
\cmidrule(lr){2-6}
\textbf{True} / \textbf{Pred.} & GEN. & FAC. & GEO. & MED. & UNSURE \\
\midrule
GEN. & 9778 & 52 & 84 & 83 & 3 \\
FAC. & 5 & 7164 & 0 & 2 & 7 \\
GEO. & 209 & 3 & 4437 & 233 & 518 \\
MED. & 144 & 409 & 313 & 16868 & 44 \\
\bottomrule
\end{tabular}

\vspace{0.5em}

\begin{tabular}{lccccc}
\toprule
& \multicolumn{5}{c}{\textbf{SC w/o CoT \& IQ}} \\
\cmidrule(lr){2-6}
\textbf{True} / \textbf{Pred.} & GEN. & FAC. & GEO. & MED. & UNSURE \\
\midrule
GEN. & 9706 & 50 & 139 & 104 & 1 \\
FAC. & 6 & 7165 & 0 & 1 & 6 \\
GEO. & 301 & 1 & 4529 & 211 & 358 \\
MED. & 190 & 275 & 412 & 16840 & 61 \\
\bottomrule
\end{tabular}
\end{table}




\end{appendices}


\section*{Declarations}

\subsection*{Availability of data and material}
All datasets used in this study are cited, publicly available, and widely used in the computer vision community. The code used for data acquisition, framework development, and evaluation of ARMDIL is available in our GitHub repository at https://github.com/Picktooth/ARMDIL-Adaptive-Router-for-Multi-Domain-Image-classification-with-LLMs.

\subsection*{Competing interests}
The authors declare that they have no competing interests.

\subsection*{Funding}
No funding was received for this study.

\subsection*{Authors' contributions}

\begin{itemize}
    \item \textbf{Daniel Perkins}: Directed project conceptualization, organization, and task delegation; conducted the literature review; developed the training infrastructure for the DINOv3 backbone; and served as the primary author of the manuscript.
    \item \textbf{John Squires:} Configured the computational environment and training pipelines; engineered the prompts and execution code for the MLLM router; and conducted the evaluation of alternative routing configurations.
    \item \textbf{Janou Milligan:} Trained and evaluated the ResNet and DINO domain experts; optimized and debugged the model training pipelines; and contributed to manuscript drafting and editing.
    \item \textbf{Chandra Raskoti:} Implemented and trained the Vision-Language Model (VLM) classifiers; and designed the architectural overview diagram (Figure \ref{fig:system_diagram}).
    \item \textbf{Linda Ungerboeck:} Developed and evaluated the majority vote and neural network router baselines.
\end{itemize}

\subsection*{Acknowledgements}
We thank Dr. Amir Sadovnik for his valuable feedback and guidance on this work. We also acknowledge the HPC resources provided by Oak Ridge National Laboratory that we used to train our domain experts.

\bibliography{sn-bibliography}

\end{document}